\documentclass[sigconf]{acmart}

\usepackage{booktabs} 
\usepackage{amsmath}
\usepackage{amssymb}
\usepackage{amsthm}
\usepackage{tikz}
\usetikzlibrary{decorations.pathreplacing}
\usepackage{tabu}
\usepackage{caption}
\usepackage{float}
\usepackage{enumitem}

\usepackage[ruled]{algorithm2e} 

\SetAlFnt{\small}
\SetAlCapFnt{\small}
\SetAlCapNameFnt{\small}
\SetAlCapHSkip{0pt}
\IncMargin{-\parindent}

\def\x{{\mathbf x}}
\def\R{{\mathbb{R}}}
\def\U{{\mathbb{U}}}

\def\rank{\text{rank}}
\newcommand{\norm}[1]{\|#1\|}
\def\O{{\mathcal{O}}}
\def\P{{\mathcal{P}}}
\def\M{{\mathcal{M}}}
\def\T{{\mathcal{T}}}
\def\H{{\mathcal{H}}}
\def\la{\lambda}
\def\si{\sigma}
\def\'{\prime}
\def\Otilde{{\widetilde{O}}}
\usepackage{todonotes}
\def\poly{\text{poly}}

\setcopyright{rightsretained}

\acmConference[KDD MLG' 17]{ACM KDD}{August 2017}{Halifax, NS, Canada} 
\acmYear{2017}
\copyrightyear{2017}

\begin{document}
\title{Fast Algorithms for Learning \\ Latent Variables in Graphical Models}

\author{Mohammadreza Soltani}
\affiliation{\institution{Iowa State University}}
\email{msoltani@iastate.edu}

\author{Chinmay Hegde}
\affiliation{\institution{Iowa State University}}
\email{chinmay@iastate.edu}

\renewcommand{\shortauthors}{B. Trovato et al.}

\begin{abstract}
We study the problem of learning latent variables in Gaussian graphical models. 
Existing methods for this problem assume that the precision matrix of the observed variables is the superposition of a sparse and a low-rank component. 
In this paper, we focus on the estimation of the low-rank component, which encodes the effect of marginalization over the latent variables. 
We introduce fast, proper learning algorithms for this problem. 
In contrast with existing approaches, our algorithms are manifestly non-convex. 
We support their efficacy via a rigorous theoretical analysis, and show that our algorithms match the best possible in terms of sample complexity, while achieving computational speed-ups over existing methods. 
We complement our theory with several numerical experiments.
\end{abstract}

\maketitle

\section{Introduction}
\label{sec:intro}

\subsection{Setup}
\label{subsec:setup}

\textit{Gaussian graphical models} are a popular tool for modeling the interaction of a collection of Gaussian random variables. In Gaussian graphical models, nodes represent random variables and edges model conditional (in)dependence among the variables
~\cite{wainwright2008graphical}. 
Over the last decade, significant efforts have been directed towards algorithms for learning \emph{sparse} graphical models. 
Mathematically, let $\Sigma^*$ denote the positive definite covariance matrix of $p$ Gaussian random variables, and let $\Theta^* = (\Sigma^*)^{-1}$ be the corresponding precision matrix. Then, $\Theta^*_{ij} = 0$ implies that the $i^{\textrm{th}}$ and $j^{\textrm{th}}$ variables are conditionally independent given all other variables and the edge $(i,j)$ does not exist in the underlying graph. The basic modeling assumption is that $\Theta^*$ is sparse, i.e., such graphs possess only a few edges. Such models have been fruitfully used in several applications including astrophysics~\cite{padmanabhan}, scene recognition~\cite{souly}, and genomic analysis~\cite{yin2013}.

Numerous algorithms for sparse graphical model learning -- both statistically as well as computationally efficient -- have been proposed in the machine learning literature~\cite{friedman2008sparse,mazumder2012graphical,banerjee2008model,hsieh2011sparse}. Unfortunately, sparsity is a simplistic first-order model and is not amenable to modeling more complex interactions. For instance, in certain scenarios, only some of the random variables are directly observed, and there could be relevant \textit{latent} interactions to which we do not directly have access.  

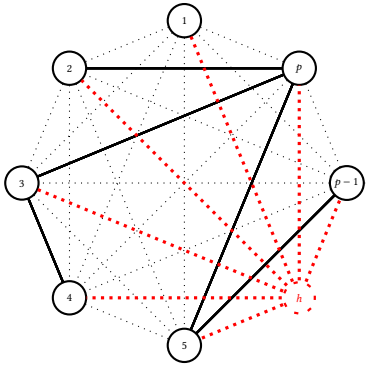
\begin{figure}[t]
\begin{center}
\begin{tikzpicture}[ultra thick,every node/.style={minimum size=3.5em},decoration={border,segment length=2mm,amplitude=0.3mm,angle=90},scale=0.4, transform shape]
  \foreach \x in {1,...,7}{%
    \pgfmathparse{(\x-1)*360/8}
    \node[draw,circle] (N-\x) at (\pgfmathresult:5.4cm) [thick] {\Large \pgfmathparse{(\x==2)?"$p$":((\x==1)?"$p-1$":int(\x-2))}\pgfmathresult};
  }
  \pgfmathparse{7*360/8}
  \node[draw,circle,red,inner sep=0.25cm,decorate] (N-8) at (\pgfmathresult:5.4cm) {\Large $h$};
  
  \foreach \x in {1,...,7}{%
    \foreach \y in {\x,...,7}{%
        \path (N-\x) edge[thin, dotted] (N-\y);
  }
  
  \path (N-2) edge[thick,-] (N-5);
  
  \path (N-2) edge[thick,-] (N-7);

  \path (N-1) edge[thick,-] (N-7);
  
  \path (N-2) edge[thick,-] (N-4);

  \path (N-5) edge[thick,-] (N-6);
  
  }
    \foreach \y in {1,...,8}{%
        \path (N-8) edge[red, very thick, dotted] (N-\y);
  }
\end{tikzpicture}
\end{center}
\caption{Illustration of effects of latent variable in graphical model learning. Solid edges represent ``true" conditional dependence, while dotted edges represent apparent dependence due to the presence of the latent variable $h$.
\label{lvgfig}}
\end{figure}

The existence of latent variables poses a significant challenge in graphical model learning since they can confound an otherwise sparse graphical model with a dense one. This scenario is illustrated in Figure~\ref{lvgfig}. Here, nodes with solid circles denote the observed variables, and solid black edges are the ``true" edges in the graphical model. One can see that the ``true" graph is rather sparse. However, if there is even a single unobserved (hidden) variable denoted by the node with the broken red circle, then it will induce dense, apparent interactions between nodes that are otherwise disconnected; these are denoted by the dotted black lines. 


A flexible and elegant method to learn latent variables in graphical models was proposed by~\cite{chandrasekaran2012latent}. At its core, the method imposes a {superposition} structure in the observed precision matrix as the sum of \emph{sparse} and \emph{low-rank} matrices, i.e.,
$ \Theta^* = S^{*} + L^{*} $.
Here, $\Theta^*, S^*, L^*$ are $p \times p$ matrices where $p$ is the number of variables. The matrix $S^*$ specifies the conditional observed precision matrix given the latent variables, while $L^*$ encodes the effect of marginalization over the latent variables. The rank of $L^*$, $r$, is equal to the number of latent variables and we assume that $r$ is much smaller than $p$.

To learn such a superposition model,~\cite{chandrasekaran2012latent} propose a regularized maximum-likelihood estimation framework, with $\ell_1$-norm and nuclear norm penalties as regularizers; these correspond to \emph{convex} relaxations of the sparsity and rank constraints, respectively. Using this framework, they prove that such graphical models can be learned with merely $n = O(pr)$ random samples. However, this statistical guarantee comes at a steep computational price; the framework involves solving a semidefinite program (SDP) with $p^2$ variables and is computationally very challenging. Several subsequent works~\cite{ma2013alternating,hsieh2014quic} have attempted to provide faster algorithms, but all known (provable) methods involve at least \emph{cubic} worst-case running time. 

\subsection{Our contributions}
\label{contri}

In this paper, we provide a new class of \emph{fast} algorithms for learning latent variables in Gaussian graphical models. Our algorithms are (i) \emph{provably statistically efficient}: they achieve the optimal sample complexity of latent variable learning; (ii) \emph{provably computationally efficient}: they are linearly convergent, and their per-iteration time is close to the best possible. 

We clarify the above claims using some notation. Suppose that we observe samples $X_1,X_2,\ldots,X_n \overset{i.i.d}{\thicksim}\mathcal{N}(0,\Sigma)$ where each $X_i\in\mathbb{R}^{p}$. Let $C= \frac{1}{n}\sum_{i=1}^{n}X_iX_i^{T}$ denote the sample covariance matrix, and $\Theta^* = (\Sigma^*)^{-1}$ denote the true precision matrix; as above, we assume that $\Theta^* = S^* + L^*$. We will exclusively function in the high-dimensional regime where $n \ll p^2$. 
Our sole focus is on learning the low-rank part from samples; i.e., we pre-suppose that the sparse part, $S^*$ is a known positive definite matrix, while the low-rank part, $L^{*}$ is unknown (with rank $r\ll p$). This models the situation where the``true" edges between the observed variables are known \emph{a priori}, but there exists latent interaction among these variables that needs to be discovered. 

We estimate $L^{*}$ in the high-dimensional regime where $n\ll p^2$ by attempting to solve a  \emph{non-convex} optimization problem, following the formulation of~\cite{han2016fast}. We label this as \emph{LVM}, short for \emph{\underline{L}atent \underline{V}ariable Gaussian Graphical \underline{M}odeling}: 
\begin{equation} \label{opt_prob} 
\begin{aligned}
& \underset{L}{\text{min}}
& & F(L) = -\log \ \det (S^{*}+L) + \langle S^{*}+L,C\rangle\\
& \text{s.t.} 
& & \rank(L)\leq r, \ L\succeq 0 .
\end{aligned}
\end{equation}



Above, $\langle.,.\rangle$ denotes the standard Frobenius inner product in matrix space, $\succeq 0$ denotes membership in the positive semi-definite (psd) cone, and the objective function $F(L)$ denotes the negative log likelihood of the samples. Problem~\eqref{opt_prob} is highly non-convex due to the rank constraint, $\rank(L)\leq r$. Moreover, the log-det loss function is highly nonlinear and challenging to handle. As mentioned earlier, most known (provable) methods solve a convex relaxation of this problem, but suffer from high computational costs. 

In contrast, we solve \eqref{opt_prob} without convex relaxation. Specifically, we propose two non-convex algorithms for solving~\eqref{opt_prob}. Our first algorithm, that we call \emph{LVM with Exact Projections}, or {EP-LVM}, performs (non-convex) projected gradient descent on the objective function $F(L)$, together with the low-rank and psd constraints. This algorithm yields sample-optimal results, but its running time can be cubic, $\Omega(p^3)$, in the number of variables. Our second algorithm, that we call \emph{LVM with Approximate Projections}, or {AP-LVM}, performs a variant of projected gradient descent with (deliberately) inaccurate projections onto the constraints. Interestingly, this algorithm also yields sample-optimal results, and its running time is \emph{nearly quadratic}, $\Otilde(p^2)$, in the dimension $p$ for a fixed number of latent variables. To the best of our knowledge, this is the fastest \emph{universal}\footnote{The running time of some other existing algorithms achieve similar scaling in $p$, but also depend adversely on matrix properties such as condition number and/or the minimum singular value.} algorithm for solving~\eqref{opt_prob}.
 
Both our proposed algorithms enjoy the following benefits:

\textbf{Sample efficiency.} For both algorithms, the sample complexity (i.e., number of samples in order to achieve a desired estimation error $\vartheta$) for learning a rank-$r$ latent variable model in $p$ variables scales as $n = O(pr)$, and this matches those of the best available methods.

\textbf{Linear convergence.} We provide rigorous analysis to show that both our proposed algorithms enjoy global linear convergence with no specific initialization step. 

\textbf{Proper learning.} Our algorithms are examples of \emph{proper} learning methods, in the sense that their output is a rank-$r$ estimate of the true latent variable model. In contrast, methods based on convex relaxation often fail to do this and return a high-rank estimate, thus potentially having a negative effect on interpretability of the discovered latent variables.

\begin{table*}[t]
\caption{Summary of our contributions, and comparison with existing methods. Here, $\gamma = \sqrt{\frac{\sigma_r}{\sigma_{r+1}}-1}$ represents the spectral gap parameter in intermediate iterations. The overall running time of the ADMM approach is marked as $\poly(p)$ since the precise rate of convergence is unknown.}
\label{Comptable}
\begin{center}
\begin{small}
\renewcommand{\arraystretch}{1.5}
\begin{tabular}{lcccr}
\hline
Algorithm & Reference & Running Time & Spectral dependency & Output rank\\
\hline
SDP  & \cite{chandrasekaran2012latent}  &$\poly(p)$  & Yes &   $\gg r$   \\
ADMM  & \cite{ma2013alternating} & $\poly(p)$ & Yes & $\gg r$ \\
QUIC \& DIRTY & \cite{yang2013dirty}  & $\Otilde(p^3)$ & Yes & $\gg r$ \\
SVP & \cite{jain2014iterative} & $\Otilde(p^3)$ & No & $\gg r$ \\
Factorized  & \cite{bhojanapalli2016dropping} & $\Otilde(p^2r / \gamma)$ & Yes & $r$ \\
EP-LVM  & \textbf{This paper} & ${\Otilde\left(p^3 \right)}$ & \textbf{No} & $\mathbf{r}$ \\
AP-LVM & \textbf{This paper} &${\Otilde(p^2 r)}$  &  \textbf{No} & $\mathbf{r}$ \\
\hline
\end{tabular}
\end{small}
\end{center}
\end{table*}

\subsection{Techniques}

Our first algorithm is a variant of the \emph{singular value projection} approach of~\cite{jain2010guaranteed}, with an extra psd projection step. Our second algorithm is a variant of approximate subspace-IHT~\cite{HegdeFastUnionNips2016} and uses a careful combination of approximate singular value decomposition techniques. While our proposed methods are structurally similar to these previously proposed methods, their analysis is considerably different; we elaborate on this below.

Our technique for establishing linear convergence of our first algorithm (EP-LVM) is based on bounding the restricted strong convexity/smoothness (RSC/RSS) constants~\cite{negahban2009unified} of the objective function $F(L)$ in~\eqref{opt_prob}. The key observation is that $F(L)$ is globally strongly convex, and when restricted to any compact psd cone, it also satisfies strong smoothness. The above analysis technique is fairly standard~\cite{jain2014iterative}. However, unlike in previously considered scenarios, the RSS/RSC constants of $F(L)$ are harder to bound. These may vary across iterations, and depend on several properties of the true precision matrix $\Theta^*$. Therefore, additional effort is required to establish global linear convergence. 

As a byproduct of this analysis, we show that with $n = O(pr)$ independent samples, EP-LVM returns an estimate up to constant error. Moreover, we show that EP-LVM provides the best empirical performance (in terms of estimation error) among all considered methods. 

However, since EP-LVM performs an exact eigenvalue decomposition (EVD) \emph{per iteration}, its overall running time can be slow since it incurs cubic per-iteration running time. Our second algorithm (AP-LVM) resolves this issue. The basic idea is to replace exact EVDs with \emph{approximate} low-rank projections in each iteration. However, it is known~\cite{hegde2015approximation} that a straightforward replacement of all EVDs with approximate low-rank projections within non-convex projected gradient descent is not a successful strategy.

In order to guarantee convergence, we use a careful combination of \emph{tail} and \emph{head} approximate low-rank projections in each iteration~\cite{HegdeFastUnionNips2016}. This enables us to improve the overall running time to $\Otilde(p^2r)$. Moreover, the statistical accuracy matches that of EP-LVM in theory (up to constants) as well as in practice (negligible loss in performance). We show that both EP-LVM and AP-LVM provide better empirical performance than convex methods.
 
Table~\ref{Comptable} provides a summary of the theoretical properties of our methods, and contrasts them with other existing methods for latent variable modeling.   

\section{Relation to Prior Work}
\label{priorworks}
Learning graphical models in high dimensional settings have been of special interest, and most existing works assume some sort of low-dimensional structure on the covariance or precision matrix~\cite{cai2016estimating}. Among all the structured models, sparse graphical model learning has received the most attention. The typical approach for learning sparse graphical models is to obtain a regularized maximum likelihood (ML) estimate given the observations. The work of~\cite{ravikumar2011high} establishes the statistical efficiency of the regularized ML estimate. Parallel to such statistical analysis is the development of efficient computational techniques for solving the regularized ML estimate~\cite{friedman2008sparse,mazumder2012graphical,rolfs2012iterative}. 

To capture latent interactions between variables, more complex structures (beyond mere sparsity) are necessary.~\cite{chandrasekaran2012latent} propose a superposition of sparse and low-rank structure in the precision matrix. 
To solve this latent variable problem, they introduce an extra nuclear norm term to the regularized ML objective function as a convex surrogate of the rank. However, they propose using a generic semi-definite programming (SDP) solver, which is very cumbersome for even moderate size problem. Subsequently, the authors in~\cite{ma2013alternating} have proposed Alternating Direction Method of Multipliers (ADMM) which scales to relatively large size problems. The work of \cite{yang2013dirty,hsieh2014quic} also consider general superposition structures in precision matrix estimation. However, the running time of these methods is still cubic in the number of variables ($p^2$). 

Problem \eqref{opt_prob} is an instance of the more general problem of \emph{low-rank matrix recovery}. Broadly, three classes of approaches for low-rank matrix recovery exist. The first (and most popular) class is based on convex relaxation~\cite{candesrecht2010,ma2013alternating,hsieh2014quic}; while their statistical properties are well-established, such methods often suffer from high computational costs.

Methods in the second class of approaches are fundamentally non-convex, and are based on the approach of~\cite{burer2003nonlinear}. In these algorithms, the psd rank-$r$ matrix $L$ is factorized as $L = UU^T$, where $U\in\R^{p\times r}$. Using this idea removes the difficulties caused by the non-convex rank constraint; however, the objective function is not convex anymore and proving convergence remains tricky. 
Nevertheless, under certain conditions, such methods succeed and have recently gained in popularity in the machine learning literature
~\cite{tu2016low,bhojanapalli2016dropping,park2016non,chen2015fast,zheng2015convergent}. In general, these methods need a careful spectral initialization that usually involves a full singular value decomposition, and their convergence depends heavily on the condition number of the input as well as other spectral properties of the true low-rank component.

The third class of methods are also non-convex. Unlike the second class, they do not factorize the optimization variable, $L$, but instead use low-rank projections within the classical gradient descent framework. This approach was introduced by~\cite{jain2010guaranteed} for matrix recovery from linear measurements, and was later modified for general M-estimation problems with well-behaved objective functions~\cite{jain2014iterative}. In principle, the approach of~\cite{jain2014iterative} can be used to solve~\eqref{opt_prob} (using a similar analysis of the RSS/RSC constants as in this paper.) However, it is an improper learning algorithm, and the rank of the estimate of $L^*$ is several times larger than the target rank. 
Moreover, each iteration is computationally expensive, since it involves computing an SVD in each iteration.

Our contributions in this paper fall under the third category. We first propose an iterative PSD projection algorithm similar to that of~\cite{jain2010guaranteed} but for the specific problem stated in~\eqref{opt_prob}. We then accelerate this algorithm (with no loss in statistical performance) using the approximate low-rank projections method of~\cite{HegdeFastUnionNips2016}.


\section{Preliminaries}
\label{prelim}
Throughout this paper, the minimum and maximum eigenvalues of the sparse matrix $S^*$ will be denoted by $S_p$ and $S_1$ respectively. We use $\|A\|_2$ and $\|A\|_F$ for spectral norm and Frobenius norm of a matrix $A$, respectively. We denote $A_r$ as the best rank-$r$ approximation (in Frobenius norm) of a given matrix $A$.
In addition, $\lambda_1(A), \lambda_p(A)$ denote the maximum and minimum eigenvalues of $A\in\R^{p\times p}$ respectively. 
For any subspace $U \subset \mathbb{R}^{p \times p}$, we denote $\P_U$ as the orthogonal projection operator onto $U$. 

Our analysis will rely upon on the following definition~\cite{negahban2009unified,jain2014iterative,yuan2013gradient}.
\begin{definition} \label{defRSCRSS}
A function $f$ satisfies the Restricted Strong Convexity (RSC) and Restricted Strong Smoothness (RSS) conditions if for all $L_1,L_2\in\mathbb{R}^{p\times p}$ such that $\rank(L_1)\leq r, \rank(L_2)\leq r$:
\begin{align}
\label{rscrss}
\frac{m_r}{2}\|L_2-L_1\|^2_F&\leq f(L_2) - f(L_1) - \langle\nabla f(L_1) , L_2-L_1\rangle \nonumber\\
&\leq\frac{M_r}{2}\|L_2-L_1\|^2_F, 
\end{align} 
where $m_r$ and $M_r$ are called the RSC and RSS constants respectively.
\end{definition}

We denote $\U_r$ as the set of all rank-$r$ matrix subspaces, i.e., subspaces of $\mathbb{R}^{p \times p}$ that are spanned by any $r$ atoms of the form $uv^T$ where $u, v \in \mathbb{R}^p$ are unit-norm vectors.

We will also employ the idea of \emph{head} and \emph{tail} projection introduced by~\cite{hegde2015approximation}, and instantiated in the context of low-rank approximation by~\cite{HegdeFastUnionNips2016}.
\begin{definition}[Approximate tail projection]
\label{taildef}
Let $c_{\T}>1$ be a constant. Then $\T:\R^{p\times p}\rightarrow\U_{r}$ is a $c_{\T}$-approximate tail projection algorithm if for all $L\in^{p\times p}$, $\T$ returns a subspace $W=\T(L)$ that satisfies: $\|L -\P_W L\|_F\leq c_{\T}\|L-L_r\|_F$. 
\end{definition}
\begin{definition}[Approximate head projection]
\label{headdef}
Let $0<c_{\H}< 1$ be a constant. Then $\H:\R^{p\times p}\rightarrow\U_{r}$ is a $c_{\H}$-approximate head projection if for all $L\in^{p\times p}$, the returned subspace $V=\H(L)$ satisfies: $\|\P_VL\|_F\geq c_{\H}\|L_r\|_F$.
\end{definition}


\section{Algorithms and Analysis}
\label{algtheory}

First, we present our projected gradient-descent algorithm to solve~\ref{opt_prob}. This algorithm provides the best sample complexity (both theoretical and empirical) among all existing approaches. Our algorithm, that we call \emph{LVM with exact projections} (EP-LVM), is described in pseudocode form as Alg~\ref{alg:svp}.
\begin{algorithm}[t]
\SetAlgoNoLine
\KwIn{Matrices $S^{*}$ and $C$, rank $r$, step size $\eta$.}
\KwOut{Estimates  $\widehat{L}$, $\widehat{\Theta} = S^{*} +  \widehat{L}$.}
\textbf{Initialization:} $L^0\leftarrow 0$, $t \leftarrow 0$\;
\Repeat{$t\leq T$}{
       $L^{t+1} = \mathcal{P}_r^+\left(L^{t} - \eta\nabla F(L^t)\right)$ \;
       $t\leftarrow t+1$\;
      }
\caption{EP-LVM}
\label{alg:svp}
\end{algorithm}

In Alg~\eqref{alg:svp}, the exact projection step, $\P_r^+(\cdot)$ denotes projection onto the space of rank-$r$ psd matrices. This is implemented through performing an exact eigenvalue decomposition (EVD) of the argument and selecting the nonnegative eigenvalues and corresponding eigenvectors~\cite{henrion2012projection}. The gradient of the objective function $F(L)$ in~\eqref{opt_prob} can be calculated as: 
\begin{align}
\label{Gradobjfunc}
\nabla F(L) = -(S^{*} + L)^{-1} + C  = -\Theta^{-1} + C.
\end{align}

Since $L$ is a low-rank matrix with rank $r$, it can be factorized as $L = UU^T$ for some $U\in\R^{p\times r}$. Hence, to calculate efficiently the inverse in~\eqref{Gradobjfunc}, we utilize the low-rank structure of $L$ by applying the Woodbury matrix identity: 
\[
(S^{*} + L)^{-1} = S^{*-1} -S^{*-1}U\left(I+U^TS^{*-1}U\right)^{-1}U^TS^{*-1}.
\]
We now provide our first main theoretical result, supporting the statistical and computational efficiency of EP-LVM. In particular, we derive an upper bound on the estimation error of the low-rank matrix at each iteration (Please see appendix for all the proofs).

\begin{theorem}[Linear convergence of EP-LVM]
\label{ExactSVD}
Assume that the objective function $F(L)$ satisfies the RSC/RSS conditions with constant $M_{3r}$ and $m_{3r}$. Let $J_t$ denote the subspace formed by the span of the column spaces of the matrices $L^t, L^{t+1}$, and $L^*$.
In addition, assume that $1\leq \frac{M_{3r}}{m_{3r}}\leq\frac{2}{\sqrt{3}}$. Choose step size as $\frac{0.5}{M_{3r}}\leq\eta\leq\frac{1.5}{m_{3r}}$. 
Then, EP-LVM outputs a sequence of estimates $L^t$ such that:
\begin{align}
\label{linconvEx}
\|L^{t+1} - L^{*}\|_F\leq \rho\|L^{t} - L^{*}\|_F + 2\eta\|\P_{J_t}\nabla F(L^{*})\|_F,
\end{align}
where $\rho = 2\sqrt{1+M_{3r}^2\eta^2 - 2m_{3r}\eta} < 1$.  
\end{theorem}

We will show below that the second term on the right hand side of this inequality is upper-bounded by an arbitrarily small constant with sufficient number of samples. Also, the first term decreases exponentially with iteration count. Overall, after $T = \mathcal{O}\left(\log_{1/\rho}\left(\frac{\|L^*\|_F}{\vartheta}\right)\right)$ iterations, we obtain an upper-bound of $O(\vartheta)$ on the total estimation error, indicating linear convergence.

Next, we provide bounds on the RSS/RSC constants of $F(L)$, justifying the assumptions made in Theorem~\ref{ExactSVD}.
\begin{theorem}[Bounding RSC/RSS constants.]
\label{RSCRSSex}
Let the number of samples scaled as $n=\O\left(\frac{1}{\delta^2}\left(\frac{\eta}{1-\rho}\right)^2rp\right)$ for some small constant $\delta>0$ and $\rho$ defined above. Also, assume that
$$S_p\leq S_1\leq\sqrt{\frac{2}{\sqrt{3}}}S_p - \left(1+\sqrt{r}\right)\|L^*\|_2- \delta.$$ 
Then, the loss function $F(L)$ satisfies RSC/RSS conditions with constants $m_{3r}$ and $M_{3r}$ that satisfy the assumptions of Theorem~\ref{linconvEx} in each iteration.
\end{theorem}

The above theorem states that convergence of our method is guaranteed when the eigenvalues of $S^*$ are roughly of the same magnitude, and large when compared to the spectral norm of $L^*$. We believe that this is merely a sufficient condition arising from our proof technique, and our numerical evidence shows that the algorithm succeeds for more general $S^*$ and $L^*$.
 
\textbf{Time complexity.} Each iteration of EP-LVM needs a full EVD, which requires cubic running time. Since the total number of iterations is logarithmic, the overall running time scales as $\Otilde(p^3)$.

For large $p$, the cubic running time of EP-LVM can be very challenging. To alleviate this issue, 
one can instead attempt to replace the full EVD in each iteration with an $\varepsilon$-\emph{approximate} low-rank psd projection; it is known that such projections can computed in $O(p^2 \log p)$ time~\cite{clarksonwoodruff}.
However, a na\"{i}ve replacement of the EVD with an $\varepsilon$-approximate low-rank projection method does not lead to algorithms with rigorous convergence guarantees.\footnote{Indeed, algorithms with only ``tail-approximate" projections can be shown to get stuck at a solution arbitrarily far from the true estimate, even with a large number of samples; see Section 2 of~\cite{hegde2015approximation}.} 

Instead, we use a combination of approximate \emph{tail} and \emph{head} projections, as suggested in~\cite{HegdeFastUnionNips2016}. The high level idea is that the use of \emph{two} inaccurate low-rank projections instead of one, if done carefully, will balance out the errors and will result in provable convergence. The full algorithm, that we call \emph{LVM with approximate projections} (AP-LVM), is described in pseudocode form as Alg.~\ref{alg:appsvp}. 
\begin{algorithm}[t]
\SetAlgoNoLine
\KwIn{Matrices $S^{*}$ and $C$, rank $r$, step size $\eta$.}
\KwOut{Estimates  $\widehat{L}$, $\widehat{\Theta} = S^{*} +  \widehat{L}$.}
\textbf{Initialization:} $L^0\leftarrow 0$, $t \leftarrow 0$\;
\Repeat{$t\leq T$}{
       $L^{t+1} = \T\left(L^{t} - \eta\H\left(\nabla F(L^t)\right)\right)$ \;
       $t\leftarrow t+1$\;
      }
\caption{AP-LVM}
\label{alg:appsvp}
\end{algorithm}

Note that we do not impose a psd projection within every iteration. If an application requires a psd matrix as the output (i.e., if proper learning is desired), then we can simply post-process the final estimate $L^T$ by retaining the nonnegative eigenvalues (and corresponding eigenvectors) through an exact EVD. We note that this EVD can be done only once, and is applied to the final output of Alg~\ref{alg:appsvp}. This is itself a rank-$r$ matrix, therefore leaving the overall asymptotic running time unchanged.

The choice of approximate low-rank projections is flexible, as long as the approximate tail and head projection guarantees are satisfied. We note that tail-approximate low-rank projection algorithms are widespread in the literature~\cite{clarksonwoodruff_old,drineas_mahoney,tygert}; however, head-approximate projection algorithms (or at least, algorithms with head guarantees) are less common. 

We focus on the randomized Block Krylov SVD method (BK-SVD) method of~\cite{musco2015randomized}. BK-SVD generates a rank-$r$ subspace approximating the top right $r$ singular vectors of a given input matrix. Moreover, for constant approximation factors, its running time is $\Otilde(p^2 r)$, \emph{independent} of any spectral properties of the input matrix. Formally, let $A\in\mathbb{R}^{n\times n}$ be a given matrix and let $A_r$ denote its best rank-$r$ approximation. Then BK-SVD generates a matrix $B = ZZ^TA$ which is the projection of $A$ onto  the column space of matrix $Z$ with orthonormal vectors $z_1,z_2,\ldots,z_r$. Moreover, with probability $99/100$, we have:
\begin{align}
\label{frogua}
\|A  - B\|_F\leq c_{\T}\|A -A_r\|_F.
\end{align}
where $c_{\T}>1$ is the tail projection constant. Equation~\eqref{frogua} is equivalent to the tail approximation guarantee according to Definition~\ref{taildef}. 

In addition to~\eqref{frogua},~\cite{musco2015randomized} also provide the so-called \emph{per vector} approximation guarantee for BK-SVD with probability $99/100$:
\begin{align*}
|u_i^TAA^Tu_i - z_iAA^Tz_i|\leq(1-c_{\H})\sigma_{r+1}^2,  
\end{align*}
where $u_i$ are the right eigenvectors of $A$ and $c_{\H} < 1$ is the head projection constant.~\cite{HegdeFastUnionNips2016} show that the above property implies the head approximation guarantee:
\[
\norm{B}_F \geq c_\H \norm{A_r}_F .
\]
Therefore, in AP-LVM we invoke the BK-SVD method for both head and tail projections\footnote{We note that since the BK-SVD algorithm is randomized while our definitions of tail and head guarantees are deterministic. Fortunately, the running time of BK-SVD depends only logarithmically on the failure probability of the algorithm, and therefore a union bound argument over the iterations of AP-LVM is required to precisely prove algorithmic correctness.}.
Using BK-SVD as the approximate low-rank projection method of choice, we now provide our second main theoretical result supporting the statistical and computational efficiency of AP-LVM.

\begin{theorem}[Linear convergence]
\label{AppSVD}
Assume that the objective function $F(L)$ satisfies the RSC/RSS conditions with constants $M_{2r}$ and $m_{2r}$.  
In addition, assume that $1\leq\frac{M_{2r}^2}{m_{2r}^2}\leq\frac{1}{1-\rho_0^2}$ where $\rho_0 = \frac{1}{1+c_{\T}}-\sqrt{1-\eta_0^2}$ and $\eta_0 = \left(c_{\H}m -\sqrt{1+M_{2r}^2-2m_{2r}}\right)$. Choose step size as $\frac{1-\rho_0^2}{M_{2r}}\leq\eta\leq\frac{1+\rho_0^2}{m_{2r}}$.
Let $V_t$ be the subspace returned by the head approximate projection $\H(\cdot)$ applied to the gradient.
Then, for any $t > 0$, AP-LVM outputs a sequence of estimates $L^t$ that satisfy:
\begin{align}
\label{linconvapp}
\|L^{t+1} - L^{*}\|_F\leq \rho_1\|L^{t} - L^{*}\|_F + \rho_2\|\P_{V_t}\nabla F(L^{*})\|_F,
\end{align}
where $\rho_1 = \left(\sqrt{1+M_{2r}^2\eta^2 - 2m_{2r}\eta}+ \sqrt{1-\eta_0^2}\right)(1+c_{\T})$ and $\rho_2 = \left(\frac{\eta_0}{\sqrt{1-\eta_0^2}} + 1\right)(1+c_{\T})$. 
\end{theorem}
 A similar calculation as before shows that AP-LVM converges after $T = \mathcal{O}\left(\log\left(\frac{\|L^*\|_F}{\vartheta}\right)\right)$ iterations.
%
AP-LVM is structurally similar to the approximate subspace-IHT algorithm of~\cite{HegdeFastUnionNips2016}. However, their proofs are specific to least-squares loss functions. On the other hand, the loss function $F(L)$ for recovering latent variables is complicated\footnote{Indeed, $F(L)$ is not well-defined everywhere, e.g. at matrices $L$ that have large negative eigenvalues.} and in general the RSS/RSC constants can vary across iterations. Therefore, considerable effort is needed to prove algorithm convergence. First, we provide conditions under which the assumption of RSC/RSS in Theorem~\ref{AppSVD} are satisfied.

\begin{theorem}[Bounding RSC/RSS constants]
\label{RSCRSSapp}
Let $n$ scaled as $n=\O\left(\frac{1}{\delta^{\'2}}\left(\frac{\rho_2}{1-\rho_1}\right)^2rp\right)$ for some small constant $\delta^{\'}>0$, with $\rho_1$ and $\rho_2$ as defined in theorem~\ref{AppSVD}. Also, assume that,
\begin{align}
\label{TrueLapp}
\|L^*\|_2\leq\frac{1}{1+\sqrt{r}}&\Biggl(\frac{S_p}{1+\sqrt{1-\rho_0^2}} \\
&- \frac{S_1\sqrt{1-\rho_0^2}}{1+\sqrt{1-\rho_0^2}} -\frac{c_3\rho_2}{1-\rho_1}\sqrt{\frac{rp}{n}}\Biggr). \nonumber
\end{align}
Finally, assume that:
$$S_p\leq S_1\leq\frac{1}{\sqrt{1-\rho_0^2}}(S_p-a^{\'}) - \left(1+\sqrt{r}\right)\|L^*\|_2- \delta^{\'}$$ 
where $0<a^{\'}\leq\left(1+\sqrt{r}\right)\|L^*\|_2 + \delta^{\'}$ for some $\delta^{\'}>0$.
Then, the loss function $F(L)$ satisfies RSC/RSS conditions with constants $m_{2r}$ and $M_{2r}$ that satisfy the assumptions of Theorem~\ref{AppSVD} in each iteration. 
\end{theorem}

Theorem~\ref{RSCRSSapp} specifies a family of true precision matrices $\Theta^* = S^* + L^*$ that can be provably estimated using our approach with an optimal number of samples. Note that since we do not perform psd projection within AP-LVM, it is possible that some of the eigenvalues of $L^t$ are negative. Next, we show that with high probability, the absolute value of the minimum eigenvalue of $L^t$ is small.

\begin{theorem}
\label{mineigval}
Under the assumptions in Theorem~\ref{RSCRSSapp} on $L^*$, if we use AP-LVM to generate a rank $r$ matrix $L^t$ for all $t=1,\ldots,T$, then with high probability the minimum eigenvalue of $L^t$ satisfies: $
\la_p(L^t)\geq -a^{\'}$ where $0<a^{\'}\leq\left(1+\sqrt{r}\right)\|L^*\|_2+ \frac{c_3\rho_2}{1-\rho_1}\sqrt{\frac{rp}{n}}$.
\end{theorem}   


\textbf{Time complexity.} Each iteration of AP-LVM needs a head and a tail projection on the rank $2r$ and rank $r$ matrices respectively. According to~\cite{musco2015randomized}, these operations  takes $k' = \O\left(\frac{p^2r\log p}{\sqrt{\varepsilon}}\right)$ for error $\varepsilon$. Since the total number of iterations is once again logarithmic, the overall running time scales as $\widetilde{O}(p^2 r)$.  
 
The above analysis shows that our proposed algorithms for discovering latent variables are linearly convergent. We now show that they converge to the true underlying low-rank matrix. The quality of the estimates in Theorems~\ref{ExactSVD} and~\ref{AppSVD} is upper-bounded by the gradient terms $\|\P_{J_t}\nabla F(L^{*})\|_F$ and $\|\P_{V_t}\nabla F(L^{*})\|_F$ in~\eqref{linconvEx} and~\eqref{linconvapp}, respectively, within each iteration. The following theorem bounds these gradient terms in terms of the number of observed samples.

\begin{theorem}
\label{BoundGrad}
Under the assumptions of Theorem~\ref{ExactSVD}, for any fixed $t$ we have:
\begin{align}
\label{BGRExc}
\|\P_{J_t}\nabla F(L^{*})\|_F\leq c_2\sqrt{\frac{rp}{n}},
\end{align}
Similarly, under the assumptions of Theorem~\ref{AppSVD}, 
\begin{align}
\label{BGRapp}
\|\P_{V_t}\nabla F(L^{*})\|_F\leq c_3\sqrt{\frac{rp}{n}},
\end{align}
Both hold with probability at least $1 - 2\exp(-p)$ where $c_2,c_3>0$ are absolute constants.
\end{theorem}

\begin{figure*}[ht]
\begin{center}
\begin{tabular}{ccc}
\includegraphics[trim = 8mm 65mm 15mm 65mm, clip, width=0.32\linewidth]{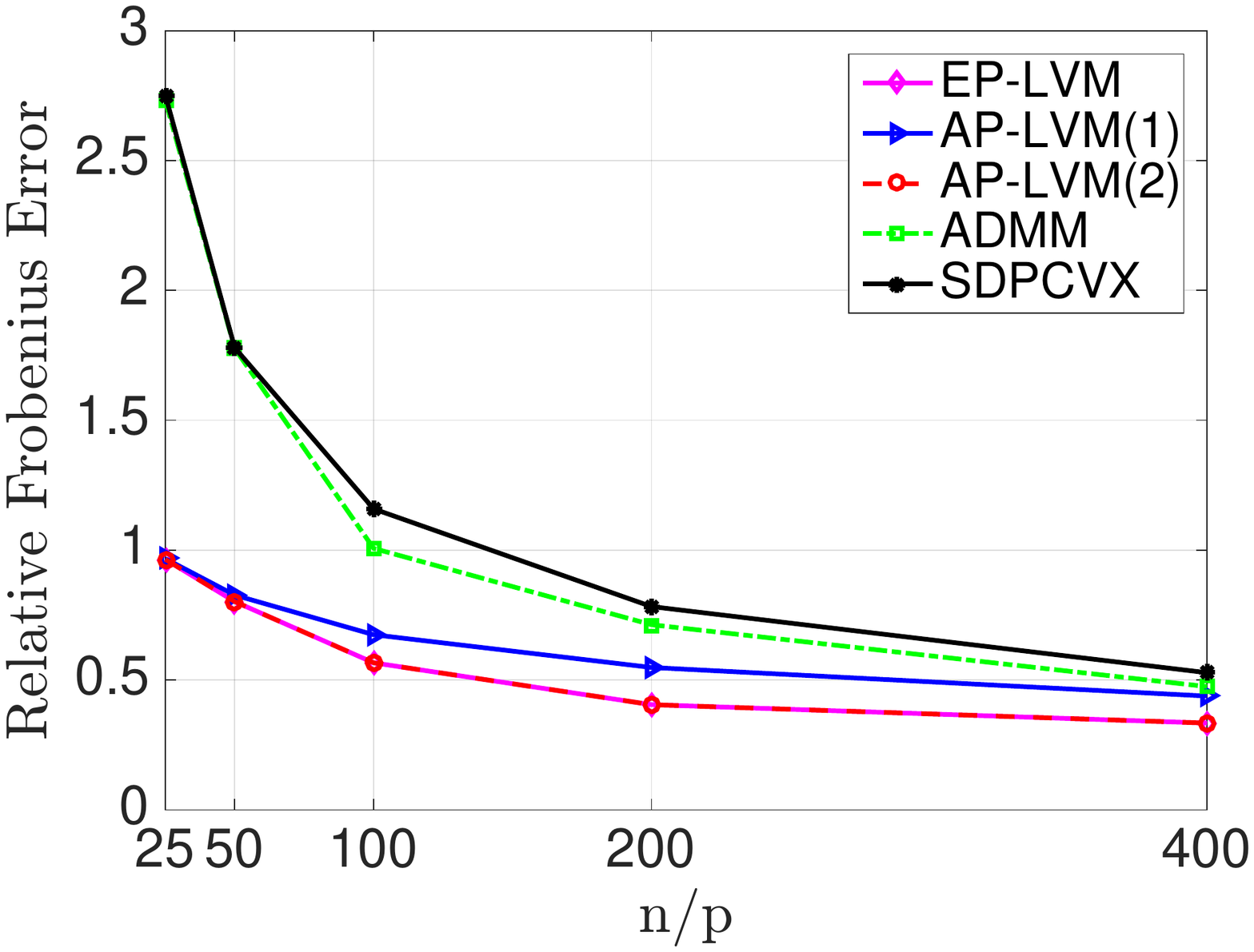} &
\includegraphics[trim = 8mm 65mm 15mm 65mm, clip, width=0.32\linewidth]{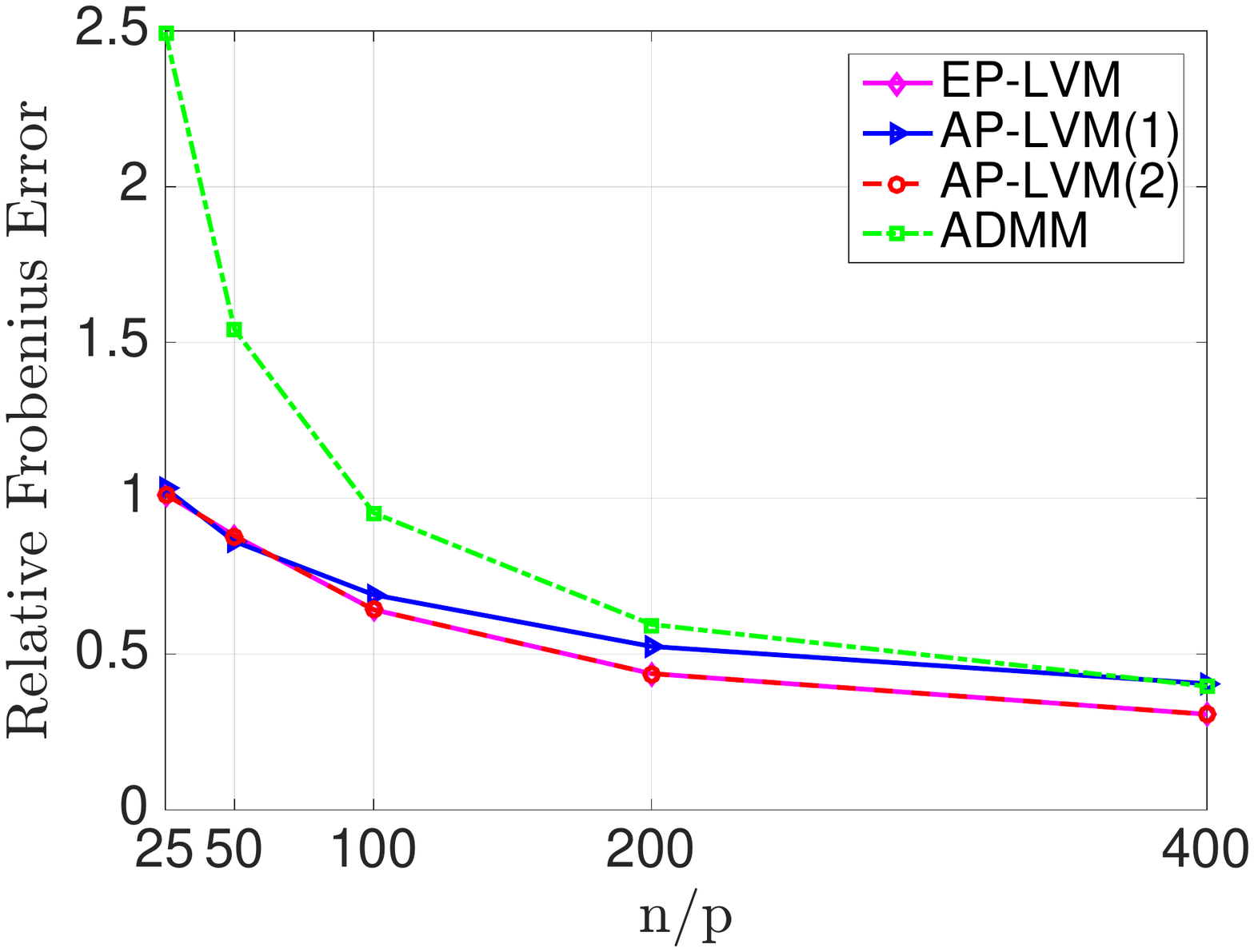} &
\includegraphics[trim = 8mm 65mm 15mm 65mm, clip, width=0.32\linewidth]{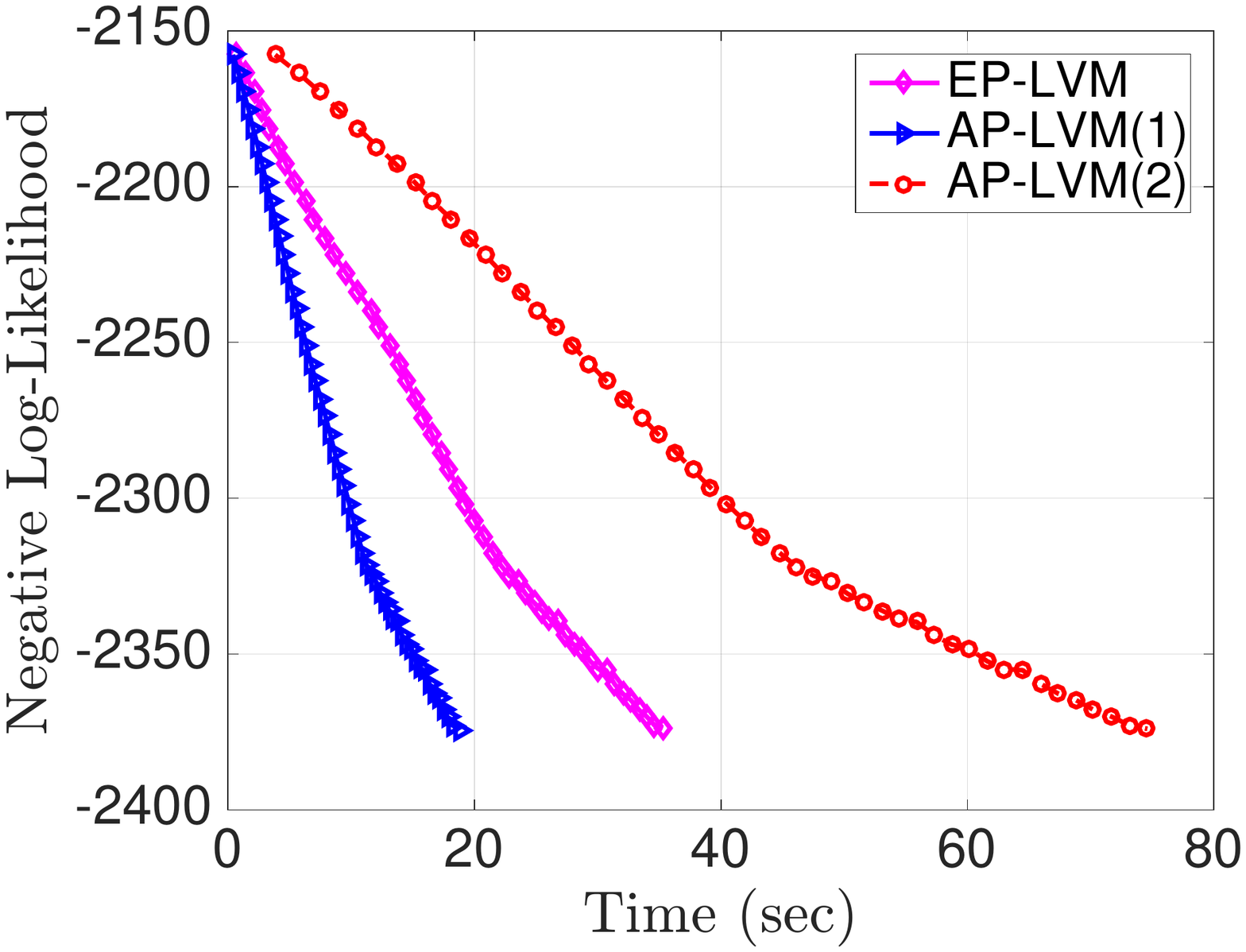}\\
$(a)$ & $(b)$ & $(c)$ 
\end{tabular}
\caption{Comparison of algorithms both in synthetic and real data. (a) relative error of $L$ in Frobenius norm with $p =100$. (b) relative error of $L$ in Frobenius norm with $p =1000$. (c) NLL versus time in Rosetta data set with $p=1000$.}
\label{p1000nllrel}
\end{center}
\end{figure*}

\textbf{Sample complexity.} Plugging in the upper bounds in~\eqref{BGRExc} and~\eqref{BGRapp} into Theorems~\ref{RSCRSSex} and ~\ref{RSCRSSapp}, the sample complexity of both algorithms scales as $n = \O(pr)$ to achieve constant estimation error. This matches the number of degrees of freedom of a $p \times p$ matrix with rank $r$.

\section{Experiments} 
\label{sec:exp}

We provide a range of numerical experiments supporting our proposed algorithms and comparing with existing convex approaches. Our comparisons is with the regularized maximum likelihood approach of~\cite{chandrasekaran2012latent}, which we solve using CVX~\citep{grant2008cvx}. 
The second algorithm that we have used is a modification of the ADMM-type method proposed by~\cite{ma2013alternating}. We assume that our algorithms are provided with the rank parameter $r$, and have manually tuned step-sizes/regularization parameters of all algorithms to achieve best possible performance.

\textbf{Synthetic data.} we use a diagonal matrix with positive values for the (known) sparse part, $S^*$. For a given number of observed variables $p$, we set $r= 5\%$ as the number of latent variables. We then follow the method proposed in~\cite{ma2013alternating} for generating the sparse and low-rank components $S^*$ and $L^*$. For simplicity, we impose the sparse component to be psd by forcing it to be diagonal. All reported results on synthetic data are the average of 5 independent Monte-Carlo trials.  
Our observations comprise $n$ samples, $x_1,x_2,\ldots,x_n \overset{i.i.d}{\thicksim}\mathcal{N}(0,(S^*+L^*)^{-1})$. In our experiments, we used a full SVD as projection step in EP-LVM. (Due to numerical stability, we use SVD rather than EVD.) For AP-LVM, we compare two versions: \emph{AP-LVM(1)} denotes the use of BK-SVD for the approximate tail and head projections, while \emph{AP-LVM(2)} denotes the use of the more well-known (but spectrum-dependent) Lanczos method for these projections. 
 
In the first experiment, we set $p=100$, $n=400p$, and $r=5$. Table~\ref{p100n400ptable} lists several metrics that we use for algorithm comparison. An algorithm terminates if it satisfies one of two conditions: the evaluated objective function in the estimated $L$ in each iteration falls below the true negative log likelihood (NLL) (i.e., $F(L^*)$), or the total number of iterations exceeds $600$. From Table~\ref{p100n400ptable}, we see that both EP-LVM, AP-LVM(1) and AP-LVM(2) produce better estimates of $L$ compared to ADMM and CVX, with AP-LVM(1) and AP-LVM(2) having the edge in running time and EP-LVM having the edge in accuracy. Note that the convex methods strictly produce an estimate of rank larger than 5 (indicating that they are \emph{improper} learning methods). As anticipated, the total running time with CVX is much larger than other algorithms. Finally, the estimated objective function for our proposed algorithms is very close to the optimal (true) objective function compared to ADMM and CVX. 

We increase the dimension to $p=1000$ and reported the same metrics in Table~\ref{p1000n400ptable} similar to Table~\ref{p100n400ptable}. Since CVX cannot solve the problem with size $p=1000$, we did not report its results. Again, we get the same conclusions as Table~\ref{p100n400ptable}; however, the improvement obtained by AP-LVM in terms of running time is considerably magnified. 

\begin{table*}
\caption{Comparison of different algorithms for $p=100$ and $n=400p$. NLL stands for negative log-likelihood.}
\label{p100n400ptable}
\vskip 0.1in
\begin{center}
\begin{small}
\begin{sc}
\begin{tabular}{lcccccr}
\hline
Alg & Estimated NLL & Estimated NLL w/reg. & True NLL & True NLL w/reg. \\
\hline
EP-LVM & $-8.884646e+01$  & $-$ & $-8.884365e+01$ & $-$ \\
AP-LVM(1)  &$-8.883135e+01$ & $-$ & $-8.884365e+01$ & $-$  \\
AP-LVM(2)  & $-8.884646e+01$ & $-$ & $-8.884365e+01$ & $-$  \\
ADMM & $-9.374270e+01$ & $-9.372003e+01$ & $-$ & $ -8.639705e+01$ & \\
CVX & $-8.891208e+01$  & $-8.889070e+01$ & $-$ & $-8.883386e+01$ \\
\hline
Alg & Relative error & Per-iteration time & Total time & Output rank\\
\hline
EP-LVM & $3.341525e-01$  & $4.982088e-03 $ & $2.386314e+00$ & $5$ \\
AP-LVM(1)  & $4.381772e-01$ & $5.626742e-03$ & $1.886186e+00$ & $5$  \\
AP-LVM(2)  & $3.341525e-01$ & $1.017454e-02$ & $4.710893e+00$ & $5$  \\
ADMM & $ 4.746382e-01$ & $9.093788e-03$ & $5.069543e+00$ & $49$\\
CVX& $5.281450e-01$ & - & $8.505811e+02$ &$100$ \\
\hline
\end{tabular}
\end{sc}
\end{small}
\end{center}
\end{table*}

\begin{table*}[ht]
\caption{Comparison of different algorithms for $p=1000$ and $n=400p$.} 
\label{p1000n400ptable}
\vskip 0.1in
\begin{center}
\begin{small}
\begin{sc}
\begin{tabular}{lcccccr}
\hline
Alg & Estimated NLL & Estimated NLL w/reg. & True NLL & True NLL w/reg. \\
\hline
EP-LVM & $-2.640322e+03$  & $-$ & $-2.640204e+03 $ & $-$ \\
AP-LVM(1) & $-2.640186e+03 $ &$-$ & $-2.640204e+03$ & $-$  \\
AP-LVM(2) & $-2.640322e+03 $ & $-$ & $-2.640204e+03$ & $-$  \\
ADMM & $-2.640565e+03$ & $-2.640000e+03$ & $-$ & $-2.522379e+03$ \\
\hline
Alg & Relative error & Per-iteration time & Total time & Output rank\\
\hline
EP-LVM & $3.065917e-01$ & $2.557906e-01$ & $1.534744e+02$ & $50$\\
AP-LVM(1) & $4.048012e-01$ & $9.880854e-02$ & $ 5.928513e+01$& $50$\\
AP-LVM(2) & $3.065917e-01$ & $3.759073e-01 $ & $2.255444e+02$& $50$\\
ADMM & $3.962763e-01$ & $5.990084e-01$ & $3.397271e+02$ & $350$ \\
\hline
\end{tabular}
\end{sc}
\end{small}
\end{center}
\vskip -0.1in
\end{table*}

\begin{table*}[!t]
\caption{Comparison of different algorithms for $p=100$ and $n=50p$. NLL stands for negative log-likelihood.}
\label{p100n50ptable}
\vskip 0.1in
\begin{center}
\begin{small}
\begin{sc}
\begin{tabular}{lcccccr}
\hline
Alg & Estimated NLL & Estimated NLL w/reg. & True NLL & True NLL w/reg. \\
\hline
EP-LVM & $-8.889947e+01$  & $-$ & $-8.887721e+01$ & $-$ \\
AP-LVM(1)  & $-8.884089e+01$ & $-$ & $-8.887721e+01$ & $-$  \\
AP-LVM(2)  & $-8.889947e+01$ & $-$ & $-8.887721e+01$ & $-$  \\
ADMM    & $-8.732370e+01$ & $-8.727177e+01$ & $-$ & $-8.643062e+01$  \\
CVX    & $-8.946498e+01$  & $-8.941170e+01$ & $-$ & $-8.886743e+01$    \\
\hline
Alg & Relative error & Per-iteration time & Total time & Output rank\\
\hline
EP-LVM & $8.020263e-01$ & $4.652518e-03$ & $2.791511e+00$ & $5$\\
AP-LVM(1)  & $8.269273e-01$ & $6.010683e-03$ & $3.387369e+00$ & $5$  \\
AP-LVM(2)  & $8.020263e-01$ & $9.625240e-03$ & $5.775144e+00$ & $3$  \\
ADMM &$1.776615e+00$ & $1.281858e-02$ & $7.691148e+00$ & $52$ \\
CVX &$1.779234e+00$ & $-$ & $8.416155e+02$ &$100$\\
\hline
\end{tabular}
\end{sc}
\end{small}
\end{center}
\end{table*}

\begin{table*}[!t]
\caption{Comparison of different algorithms for $p=1000$ and $n=50p$. 
}
\label{p1000n50ptable}
\vskip 0.1in
\begin{center}
\begin{small}
\begin{sc}
\begin{tabular}{lcccccr}
\hline
Alg & Estimated NLL & Estimated NLL w/reg. & True NLL & True NLL w/reg. \\
\hline
EP-LVM & $-2.640797e+03 $  & $-$ & $-2.640199e+03$ & $-$ \\
AP-LVM(1) & $-2.640143e+03 $ & $-$ & $-2.640199e+03$ & $-$  \\
AP-LVM(2) & $-2.640797e+03 $ & $-$ & $-2.640199e+03$ & $-$  \\
ADMM & $-2.645466e+03$ & $-2.643407e+03$ & $-$ & $-2.522374e+03$ \\
\hline
Alg & Relative error & Per-iteration time & Total time & Output rank\\
\hline
EP-LVGGM & $8.789615e-01$ & $2.510983e-01$ & $1.506590e+02$ & $50$\\
AP-LVGGM(1) & $8.609531e-01$ & $9.864577e-02$ & $5.918746e+01$& $50$\\
AP-LVGGM(2) & $8.789615e-01$ & $3.756105e-01$ & $2.253663e+02$& $50$\\
ADMM & $1.540313e+00$ & $6.669523e-01$ & $3.793097e+02$ & $462$ \\
\hline
\end{tabular}
\end{sc}
\end{small}
\end{center}
\vskip -0.1in
\end{table*}

In addition, Tables~\ref{p100n50ptable} and~\ref{p1000n50ptable} shows the same experiment discussed in Tables~\ref{p100n400ptable} and~\ref{p1000n400ptable} but for small number of samples, $n=50p$.

In Figures~\ref{p1000nllrel} (a) and (b), we graphically compare four algorithms in terms of the relative error of the estimated $L$ in Frobenius norm versus the ``oversampling" ratio $n/p$. In this experiment, we fixed $p=100$ in (a) and $p=1000$ in (b) and vary $n$. We observe that EP-LVM, AP-LVM(1), and AP-LVM(2) estimate the low-rank matrix even for the regime where $n$ is very small, whereas both ADMM and CVX does not produce very meaningful results. 

\textbf{Real data.} We evaluate our methods through the \emph{Rosetta} gene expression data set~\cite{hughes2000functional}. This data set includes 301 samples with 6316 variables. We run the ADMM algorithm by~\cite{ma2013alternating} with $p=1000$ variables and obtained an estimate of the sparse component $S^*$. Then we used $S^*$ as the input for EP-LVM, AP-LVM(1) and AP-LVM(2). The target rank for all three algorithms is set to be the same as that returned by ADMM.   
In Figure~\ref{p1000nllrel} plot (c), we illustrate the NLL for these three algorithms versus wall-clock time (in seconds) over 50 iterations. We observe that all three algorithms demonstrate linear convergence, as predicted in the theory. Among the three algorithms, AP-LVM(1) obtains the quickest rate of decrease of the objective function. 


\section*{Acknowledgements}
This work was supported in part by grants from the National Science Foundation and NVIDIA.

\bibliographystyle{alpha}
\bibliography{../Common/mrsbiblio,../Common/chinbiblio,../Common/csbib} 

\flushbottom
\onecolumn
\section{Appendix}
\label{append}


We provide full proofs of all theorems discussed in this paper. 

Below, the expression $C+D$ for two sets $C$ and $D$ refers to the \textit{Minkowski} sum of two sets, defined as $C+D = \{c+d \ | \ c\in C, \ d\in D\}$ for given sets $C$ and $D$. Also, $\M(\U_r)$ denotes the set of vectors associated with $\U_r$, the set of all rank-r matrix subspaces. Furthermore $\sigma_i(A)$ denotes the $i^{th}$ largest singular value of matrix $A$.   
We need the following equivalent definitions of restricted strongly convex and restricted strong smoothness conditions.

\begin{definition} \label{defRSCRSS_app}
A function $f$ satisfies the Restricted Strong Convexity (RSC) and Restricted Strong Smoothness (RSS) conditions if one of the following equivalent definitions is satisfied for all $L_1,L_2, L\in\mathbb{R}^{p\times p}$ such that $\rank(L_1)\leq r, \rank(L_2)\leq r,rank(L)\leq r$:
\begin{align}
&\frac{m_r}{2}\|L_2-L_1\|^2_F \leq f(L_2) - f(L_1) - \langle\nabla f(L_1) , L_2-L_1\rangle\leq\frac{M_{r}}{2}\|L_2-L_1\|^2_F, \label{rscrss_app1}\\
&\hspace{5mm}\|L_2-L_1\|^2_F\leq\langle\P_U\left(\nabla f(L_2) - \nabla f(L_1)\right), L_2-L_1\rangle \leq M_{r}\|L_2-L_1\|^2_F, \label{rscrss_app2}\\
&\hspace{28mm}m_{r} \leq\|\P_U\nabla^2 f(L)\|_2 \leq M_{r},\label{rscrss_app3} \\
&\hspace{5mm}m_{r}\|L_2-L_1\|_F \leq\|\P_U\left(\nabla f(L_2) - \nabla f(L_1)\right)\|_F \leq M_{r}\|L_2-L_1\|_F,\label{rscrss_app4}
\end{align} 
where $U$ is the span of the union of column spaces of the matrices $L_1$ and $L_2$. Here, $m_r$ and $M_r$ are the RSC and RSS constants, respectively. 
\end{definition}
The key observation is that the objective function in~\eqref{opt_prob} is globally strongly convex, and when restricted to any compact psd cone, it also satisfies the smoothness condition. As a result, it satisfies RSC/RSS conditions.

\begin{proof}[Proof of Theorem~\ref{ExactSVD}]
Let $V^t, V^{t+1}$, and $V^{*}$ denote the bases for the column space of $L^t, L^{t+1}$, and $L^*$, respectively. By definition of set $J$ in the theorem, $V^t\cup V^{t+1}\cup V^*\subseteq J_t := J$. Define $b = L^t -\eta\P_J\nabla F(L^t)$. We have:
\begin{align}
\label{linExact}
\|L^{t+1} - L^*\|_F&\leq\|L^{t+1} - b\|_F + \|b-L^*\|_F \\
& \overset{e_1}{\leq}2\|b-L^*\|_F \nonumber \\
& \leq2\|L^t - L^*-\eta\P_J\nabla F(L^t)\|_F \nonumber \\
&\overset{e_2}{\leq}2\|L^t - L^*-\eta\P_J\left(\nabla F(L^t)- \nabla F(L^*) \right)\|_F + 2\eta\|\P_J\nabla F(L^*)\|_F \nonumber \\
&\overset{e_3}{\leq}2\sqrt{1+M_{3r}^2\eta^2 - 2m_{3r}\eta}\|L^t-L^*\|_F + 2\eta\|\P_J\nabla F(L^*)\|_F 
\end{align}
where $e_1$ holds since $L^{t+1}$ is generated by projecting onto the set of matrices with rank $r$ and retaining only the positive eigenvalues; and by definition of $J$, $L^{t+1}$ also has the minimum Euclidean distance to $b$ over all matrices with rank $r$. Moreover, $e_2$ holds by applying triangle inequality and $e_3$ is obtained by combining by lower bound in~\eqref{rscrss_app2} and upper bound in~\eqref{rscrss_app4}, i.e.,
$$\|L^t - L^* - \eta^{\prime}\left(\nabla_J F(L^t) -\nabla_J F(L^*)\right)\|_2^2\leq(1+{\eta^{\prime}}^2M_{3r}^2-2\eta^{\prime} m_{3r})\|L^t-L^*\|_2^2.$$
For~\eqref{linExact} to imply convergence, we require that $\sqrt{1+M_{3r}^2\eta^2 - 2m_{3r}\eta}<\frac{1}{2}$. By solving this quadratic inequality with respect to $\eta$, we obtain the conditions $1\leq \frac{M_{3r}}{m_{3r}}\leq\frac{2}{\sqrt{3}}$, and $\frac{0.5}{M_{3r}}\leq\eta\leq\frac{1.5}{m_{3r}}$.
If we initialize at $L^0 = 0$, then we obtain $\vartheta$ accuracy after $T = \mathcal{O}\left(\log\left(\frac{\|L^*\|_F}{\vartheta}\right)\right)$.
\end{proof}

\begin{proof}[Proof of Theorem~\ref{AppSVD}]
Define $b^{\prime} = L^t -\eta\H\nabla F(L^t)$. Let $Y\in\U_{2r}$, $W = \T({b^\prime})$, and $V := V_t = \H(\nabla F(L^t))$. Also, by the definition of the tail projection, we have $L^{t}\in\M(\U_{r})$. Hence, we have:
\begin{align}
\label{headtailproof}
\|L^{t+1} - L^*\|_F &=\| L^* - \P_W(b^{\prime})\|_F \nonumber \\
&\leq\|L^* - b^{\prime}\|_F + \|b^{\prime} - \P_W(b^{\prime})\|_F \nonumber \\
&\overset{e_1}{\leq}(1+c_{\T})\|b^{\prime} - L^*\|_F  \nonumber \\
&\overset{}{=}(1+c_{\T})\|L^t - L^* - \eta\H\left(\nabla F(L^t\right))\|_F \nonumber \\
&\overset{e_2}{=}(1+c_{\T})\|L^t - L^* - \eta\P_V\nabla F(L^t)\|_F \nonumber \\
&\overset{e_3}{\leq}(1+c_{\T})\|\P_V(L^t-L^*) + \P_{V^{\bot}}(L^t-L^*) - \eta\P_V\nabla F(L^t)\|_F \nonumber \\
&\overset{}{\leq}(1+c_{\T})\|\P_V(L^t-L^*) - \eta\P_V\nabla F(L^t)\|_F + (1+c_{\T})\|\P_{V^{\bot}}(L^t-L^*)\|_F  \nonumber \\
&\overset{e_4}{\leq}(1+c_{\T})\|L^t-L^* - \eta\P_{V+Y}\left(\nabla F(L^t)-\nabla F(L^*)\right)\|_F \nonumber\\
& \quad\quad\quad\quad\quad + (1+c_{\T})\|\P_{V^{\bot}}(L^t-L^*)\|_F  + (1+c_{\T})\|\P_V\nabla F(L^*)\|_F 
\end{align}
In the above inequalities, $e_1$ is due to the triangle inequality and the definition of approximate tail projection, $e_2$ is obtained by the definition of approximate head projection, $e_3$ holds by decomposing of the residual $L^t - L^*$ in the subspace $V$ and $V^{\bot}$, and finally $e_4$ is due to the triangle inequality and the fact that $L^t - L^*\in\M(\U_{2r})$ and $V\subseteq V+Y$. 

As we can see in~\eqref{headtailproof}, we have three terms that we need to bound. For the first term we have:
\begin{align}
\label{firsApp}
(1+c_{\T})\|L^t-L^* - \eta\P_{V+Y}\left(\nabla F(L^t)-\nabla F(L^*)\right)\|_F\leq(1+c_{\T})\sqrt{1+M_{2r}^2\eta^2 - 2m_{2r}\eta}\|L^t-L^*\|_F,  
\end{align}
where the above inequality holds due to the RSC/RSS assumption on the objective function, $F(L)$ similar to $e_3$ in~\eqref{linExact}. The third term in~\eqref{headtailproof} is bounded by the argument given in~\eqref{BGRapp} (see section~\ref{algtheory}). To bound the second term, $\|\P_{V^{\bot}}(L^t-L^*)\|_F$, we follow the proof technique in~\cite{HegdeFastUnionNips2016}. First we have: 
\begin{align}
\label{projonVlow}
\|\P_V\nabla F(L^t)\|_F&\geq c_{\H}\|\P_Y\nabla F(L^t)\|_F\nonumber \\
&\overset{e_1}{\geq}c_{\H}\|\P_Y\left(\nabla F(L^t) -\nabla F(L^*)\right)\|_F - c_{\H}\|\P_Y\nabla F(L^*)\|_F \nonumber \\
&\overset{e_2}{\geq}c_{\H}m\|L^t - L^*\|_F - c_{\H}\|\P_Y\nabla F(L^*)\|_F,
\end{align}
where $e_1$ is followed by adding and subtracting $\|\P_Y\nabla F(L^*)\|_F$, and then invoking the triangle inequality. Also, $e_2$ holds due to the lower bound in the definition of $(14)$ in the RSC/RSS conditions. In addition, we can bound $\|\P_V\nabla F(L^t)\|_F$ from the above as follows:
\begin{align}
\label{projonVup}
\|\P_V\nabla F(L^t)\|_F&\overset{e_1}{\leq}\|\P_V\left(\nabla F(L^t) -\nabla F(L^*)\right) - \P_V(L^t- L^*)\|_F +\|\P_V(L^t- L^*)\|_F + \|\P_V\nabla F(L^*)\|_F\nonumber \\
&\overset{e_2}{\leq}\|\P_{V+Y}\left(\nabla F(L^t) -\nabla F(L^*)\right) - \P_{V+Y}(L^t- L^*)\|_F  +\|\P_V(L^t- L^*)\|_F + \|\P_V\nabla F(L^*) \|_F\nonumber \\
&\overset{e_2}{\leq}\| L^t- L^* - \P_{V+Y}\left(\nabla F(L^t) -\nabla F(L^*)\right)\|_F +  \|\P_{V}(L^t- L^*)\|_F + \|\P_V\nabla F(L^*) \|_F\nonumber \\
&\overset{e_4}{\leq}\sqrt{1+M_{2r}^2-2m_{2r}}\|L^t-L^*\|_F + \|\P_{V}(L^t- L^*)\|_F + \|\P_V\nabla F(L^*)\|_F,
\end{align}
where $e_1$ holds by adding and subtracting  $\P_V\nabla F(L^*)$ and $\P_V(L^t - L^*)$ and using triangle inequality. $e_2$ is followed by the fact that $V\subseteq V+Y$ which implies that projecting onto the extended subspace $V+Y$ instead of $V$ cannot decrease the norm. Also, $e_3$ holds since $L^t - L^*\in\M(\U_{2r})$. Finally $e_4$ holds by using RSC/RSS assumption on the objective function, $F(L)$ similar to $e_3$ in~\eqref{linExact}. As a result we have from~\eqref{projonVlow} and~\eqref{projonVup}: 
\begin{align}
\|\P_{V}(L^t- L^*)\|_F\geq\left(c_{\H}m -\sqrt{1+M_{2r}^2-2m_{2r}}\right)\|L^t-L^*\|_F - (1+c_{\H} )\|\P_V\nabla F(L^*)\|_F
\end{align}
Now we can bound the second term in~\eqref{headtailproof}, $(1+c_{\T})\|\P_{V^{\bot}}(L^t-L^*)\|_F$ since from the Pythagoras theorem, we have $\|\P_{V^{\bot}}(L^t-L^*)\|_F^2 = \|L^t-L^*\|_F^2 - \|\P_V(L^t-L^*)\|_F^2$. To do this, we invoke Claim $(14)$ in~\cite{HegdeFastUnionNips2016} which gives:
\begin{align}
\label{perPeprojVres}
(1+c_{\T})\|\P_{V^{\bot}}(L^t-L^*)\|_F\leq(1+c_{\T})\sqrt{1-\eta_0^2}\|L^t-L^*\|_F + \frac{\eta_0(1+c_{\T})}{\sqrt{1-\eta_0^2}}\|\P_V\nabla F(L^*)\|_F
\end{align}
where $\eta_0 = \left(c_{\H}m -\sqrt{1+M_{2r}^2-2m_{2r}}\right)$. We obtained the claimed bound in the theorem by combining upper bounds in~\eqref{firsApp} and~\eqref{perPeprojVres}:
\begin{align}
\label{linapp}
\|L^{t+1} - L^{*}\|_F\leq \rho_1\|L^{t} - L^{*}\|_F + \rho_2\|\P_V\nabla F(L^{*})\|_F,
\end{align}
where $\rho_1 = \left(\sqrt{1+M_{2r}^2\eta^2 - 2m_{2r}\eta}+ \sqrt{1-\eta_0^2}\right)(1+c_{\T})$ and $\rho_2 = \left(\frac{\eta_0}{\sqrt{1-\eta_0^2}} + 1\right)(1+c_{\T})$. Now to have meaningful bound in~\eqref{linapp}, we need to have $\rho_1 <1$ or $M_{2r}^2\eta^2 - 2m_{2r}\eta+1-0.25\left(\frac{1}{1+c_{\T}}-\sqrt{1-\eta_0^2}\right)^2 <0$ which implies $1\leq\frac{M_{2r}^2}{m_{2r}^2}\leq\frac{1}{1-\rho_0^2}$ where $\rho_0 = \frac{1}{1+c_{\T}}-\sqrt{1-\eta_0^2}$. Now by induction and zero initialization, we obtain the $\vartheta$ accuracy after $T = \mathcal{O}\left(\log\left(\frac{\|L^*\|_F}{\vartheta}\right)\right)$.
\end{proof}

We still need to show that why the assumptions on the RSC/RSS constants of $F(L)$ is satisfied at \emph{each} iteration of EP-LVM and AP-LVM.  
To do this, we prove Theorems~\ref{RSCRSSex} and~\ref{RSCRSSapp}. Our strategy is to establish upper and lower bounds on the spectrum of the sequence of estimates $L^t$ independent of $t$. We use the following lemma.
\begin{lemma}\cite{yuan2013gradient,boyd2004convex}
\label{hessian}
The Hessian of the objective function $F(L)$ is given by $\nabla^2F(L) = \Theta^{-1}\otimes\Theta^{-1}$ where $\otimes$ denotes the Kronecker product and $\Theta = S^* + L$. In addition if $\alpha I\preceq\Theta\preceq\beta I$ for some $\alpha$ and $\beta$, then $\frac{1}{\beta^2} I\preceq\nabla^2F(L)\preceq\frac{1}{\alpha^2} I$.
\end{lemma}
\begin{lemma}[Weyl type inequality]
\label{weyl}
For any two matrices $A,B\in\R^{p\times p}$, we have:
$$\max_{1\leq i\leq p} |\sigma_i(A+B) - \sigma_i(A)|\leq\|B\|_2.$$
\end{lemma}

If we establish an universal upper bound and lower bound on $\lambda_1(\Theta^t)$ and $\lambda_p(\Theta^t) \ \forall t=1\dots T$, then we can bound the RSC constant as $m_{}\geq\frac{1}{\lambda_1(\Theta^t)^2}$ and the RSS-constant as $M_{}\leq\frac{1}{\lambda_p(\Theta^t)^2}$ using Lemma~\ref{hessian} and the definition of RSS/RSC.

\begin{proof}[Proof of Theorem~\ref{RSCRSSex}]
Recall that by Theorem~\ref{ExactSVD}, we have $\|L^{t} - L^{*}\|_F\leq \rho\|L^{t-1} - L^{*}\|_F + 2\eta\|\P_J\nabla F(L^{*})\|_F,$ where $\rho <1$ is defined as $\rho = 2\sqrt{1+M_{3r}^2\eta^2 - 2m_{3r}\eta}$ for $1\leq \frac{M_{3r}}{m_{3r}}\leq\frac{2}{\sqrt{3}}$. By Theorem~\ref{BoundGrad}, the second term on the right hand side can be bounded by $O(\sqrt{\frac{rp}{n}})$ with high probability. Therefore, recursively applying this inequality to $L^t$ (and initializing with zero), we obtain:
\begin{align}
\label{firstexactupp}
\|L^{t} - L^{*}\|_F\leq\rho^t\|L^*\|_F+ \frac{2\eta}{1-\rho} c_2 \sqrt{\frac{rp}{n}}.
\end{align} 
Since $\rho<1$, then $ \rho^t<1$. On the other hand $\|L^*\|_F\leq\sqrt{r}\|L^*\|_2$. Hence, $\rho^t\|L^*\|_F\leq\sqrt{r}\|L^*\|_2$. Also, by the Weyl inequality, we have:
\begin{align}
\|L^t\|_2 - \|L^*\|_2\leq\|L^t-L^*\|_2\leq\|L^t-L^*\|_F.
\end{align}
Combining~\eqref{firstexactupp} and~\eqref{secondexactupp} and using the fact that $\la_1(L^{t+1})\leq\si_1(L^{t+1})$,  
\begin{align}
\la_1(L^{t})&\leq\|L^*\|_2+\|L^t-L^*\|_F\nonumber  \\
&\leq\|L^*\|_2+\sqrt{r}\|L^*\|_2+ \frac{2\eta}{1-\rho}c_2\sqrt{\frac{rp}{n}}\nonumber  
\end{align}
Hence for all $t$, 
\begin{align}
\label{secondexactupp}
\la_1(\Theta^t) = S_1+\la_1(L^t)\leq S_1 +\left(1+\sqrt{r}\right)\|L^*\|_2+ \frac{2c_2\eta}{1-\rho}\sqrt{\frac{rp}{n}}.
\end{align}
For the lower bound, we trivially have for all $t$:
\begin{align}
\la_p(\Theta^t) = \la_p(S^* +  L^t)\geq S_p.
\end{align}
If we select $n=\O\left(\frac{1}{\delta^2}\left(\frac{\eta}{1-\rho}\right)^2rp\right)$ for some small constant $\delta>0$, then~\eqref{secondexactupp} becomes:
\begin{align*}
\la_1(\Theta^t)\leq S_1 +\left(1+\sqrt{r}\right)\|L^*\|_2+ \delta.
\end{align*}
As mentioned above, we set $m_{3r}\geq\frac{1}{\lambda_1^2(\Theta^t)}$ and $M_{3r}\leq\frac{1}{\lambda_p^2(\Theta^t)}$ which implies $\frac{M_{3r}}{m_{3r}}\leq\frac{\la_1^2(\Theta^t)}{\la_p^2(\Theta^t)}$. In order to satisfy the assumption on the RSC/RSS in theorem~\ref{ExactSVD}, i.e., $\frac{M_{3r}}{m_{3r}}\leq\frac{2}{\sqrt{3}}$, we need to establish a regime such that $\frac{\la_1^2(\Theta^t)}{\la_p^2(\Theta^t)}\leq\frac{2}{\sqrt{3}}$. 
As a result, to satisfy the inequality $\frac{\la_1^2(\Theta^t)}{\la_p^2(\Theta^t)}\leq\frac{2}{\sqrt{3}}$, we need to have the following condition: 
\begin{align}
S_p\leq S_1\leq\sqrt{\frac{2}{\sqrt{3}}}S_p - \left(1+\sqrt{r}\right)\|L^*\|_2- \delta.
\end{align}
\end{proof}

\begin{proof}[Proof of Theorem~\ref{RSCRSSapp}]
The proof is similar to the proof of theorem~\ref{RSCRSSex}.  Recall that by theorem~\ref{AppSVD}, we have 
$$\|L^{t} - L^{*}\|_F\leq \rho_1\|L^{t-1} - L^{*}\|_F + \rho_2\|\P_{V_t}\nabla F(L^{*})\|_F,$$ 
where $\rho_1 = \left(\sqrt{1+M_{2r}^2\eta^2 - 2m_{2r}\eta}+ \sqrt{1-\eta_0^2}\right)(1+c_{\T})$, $\rho_2 = \left(\frac{\eta_0}{\sqrt{1-\eta_0^2}} + 1\right)(1+c_{\T})$, and the set $V_t$ is as defined in Theorem~\ref{AppSVD}. Again, by Theorem~\ref{BoundGrad}, the second term on the right hand side is bounded by $O(\sqrt{rp/n})$ with high probability. As above, recursively applying this inequality to $L^t$ and using zero initialization, we obtain:
\begin{align*}
\|L^{t} - L^{*}\|_F\leq\rho_1^t\|L^*\|_F  + \frac{\rho_2}{1-\rho_1} c_3 \sqrt{\frac{rp}{n}}.
\end{align*}
Since $\rho_1<1$, then $\rho_1^t<1$. Now similar to the exact algorithm, $\|L^*\|_F\leq\sqrt{r}\|L^*\
|_2$ and $\rho_1^t\|L^*\|_F\leq\sqrt{r}\|L^*\|_2$. , Hence with high probability,
\begin{align}
\label{secontineqApp}
\la_1(L^{t})&\leq\|L^*\|_2+\|L^t-L^*\|_F\nonumber  \\
&\leq\|L^*\|_2 + \sqrt{r}\|L^*\|_2 + \frac{\rho_2}{1-\rho_1} \sqrt{\frac{rp}{n}} \nonumber  \\
&\overset{e_1}{\leq}\left(1+\sqrt{r}\right)\|L^*\|_2+ \frac{c_3\rho_2}{1-\rho_1}\sqrt{\frac{rp}{n}},
\end{align} 
where $e_1$ holds due to~\eqref{BGRapp}. Hence, for all $t$:
\begin{align}
\label{firstineqApp}
\la_1(\Theta^t) = S_1+\la_1(L^t)\leq S_1 + \left(1+\sqrt{r}\right)\|L^*\|_2+ \frac{c_3\rho_2}{1-\rho_1}\sqrt{\frac{rp}{n}},
\end{align}

Also, we trivially have:
\begin{align}
\la_p(\Theta^t)=\la_p(S^* +  L^t)\geq S_p - a^{\prime}, \ \forall t.
\end{align}
By selecting $n=\O\left(\frac{1}{\delta^{\'2}}\left(\frac{\rho_2}{1-\rho_1}\right)^2rp\right)$ for some small constant $\delta^{\'}>0$, we can write~\eqref{firstineqApp} as follows:
\begin{align*}
\la_1(\Theta^t)\leq S_1 + \left(1+\sqrt{r}\right)\|L^*\|_2+\delta^{\'},
\end{align*}
In order to satisfy the assumptions in Theorem~\ref{AppSVD}, i.e., $\frac{M_{2r}^2}{m_{2r}^2}\leq\frac{1}{1-\rho_0^2}$ where $\rho_0 = \frac{1}{1+c_{\T}}-\sqrt{1-\eta_0^2}$ and $\eta_0 = \left(c_{\H}m -\sqrt{1+M_{2r}^2-2m_{2r}}\right)$, we need to guarantee that $\frac{\la_1(\Theta^t)}{\la_p(\Theta^t)}\leq\frac{1}{\sqrt{1-\rho_0^2}}$. 
As a result, to satisfy the inequality $\frac{\la_1(\Theta^t)}{\la_p(\Theta^t)}\leq\frac{1}{\sqrt{1-\rho_0^2}}$, we need to have the following condition on $S_1$ and $S_p$: 
\begin{align}
S_p\leq S_1&\leq\frac{1}{\sqrt{1-\rho_0^2}}(S_p-a^{\'}) - \left(1+\sqrt{r}\right)\|L^*\|_2- \delta^{\'}.
\end{align}
\end{proof}


\begin{proof}[Proof of Theorem~\ref{mineigval}]
Recall from~\eqref{secontineqApp} that with very high probability, $\|L^t\|_2{\leq}\left(1+\sqrt{r}\right)\|L^*\|_2+ \frac{c_3\rho_2}{1-\rho_1}\sqrt{\frac{rp}{n}}$. Also, we always have: $\la_p(L^t)\geq-\|L^t\|_2$. As a result:
\begin{align}
\la_p(L^t)\geq-\left(1+\sqrt{r}\right)\|L^*\|_2- \frac{c_3\rho_2}{1-\rho_1}\sqrt{\frac{rp}{n}}.
\end{align}
Now if the inequality $\left(1+\sqrt{r}\right)\|L^*\|_2+ \frac{c_3\rho_2}{1-\rho_1}\sqrt{\frac{rp}{n}}< S_p$ is satisfied, then we can select $0<a^{\'}\leq\left(1+\sqrt{r}\right)\|L^*\|_2+ \frac{c_3\rho_2}{1-\rho_1}\sqrt{\frac{rp}{n}}$. The former inequality is satisfied 
by the assumption of Theorem~\ref{RSCRSSapp} on $\|L^*\|_2$, i.e., 
$$\|L^*\|_2\leq\frac{1}{1+\sqrt{r}}\left(\frac{S_p}{1+\sqrt{1-\rho_0^2}} - \frac{S_1\sqrt{1-\rho_0^2}}{1+\sqrt{1-\rho_0^2}} -\frac{c_3\rho_2}{1-\rho_1}\sqrt{\frac{rp}{n}}\right).$$
\end{proof}

\begin{proof}[Proof of Theorem~\ref{BoundGrad}]
The proof of this theorem is a direct application of the Lemma 5.4 in \cite{Venkat2009sparse} and we restate it for completeness:
\begin{lemma}
\label{boundSmCo}
Let $C$ denote the sample covariance matrix, then with probability at least $1 - 2\exp(-p)$ we have $\|C - (S^* + L^*)^{-1}\|_2\leq c_1\sqrt{\frac{p}{n}}$ where $c_1>0$ is a constant.
\end{lemma}
By noting that $\nabla F(L^{*}) = C - (S^* + L^*)^{-1}$ and $\rank(J_t)\leq 3r$, we can bound the term on the right hand side in Theorem~\ref{ExactSVD} as:
$$\|\P_{J_t}\nabla F(L^{*})\|_F\leq \sqrt{3r}\|\nabla F(L^{*})\|_2\leq c_2\sqrt{\frac{rp}{n}} .$$
The proof for upper-bounding $\|\P_{V_t}\nabla F(L^{*})\|_F$ in Theorem~\ref{AppSVD} follows analogously. 
\end{proof}

\end{document}